\documentclass[10pt,twocolumn,letterpaper]{article}

\usepackage[pagenumbers]{cvpr} 

\newcommand{\red}[1]{{\color{red}#1}}

\definecolor{cvprblue}{rgb}{0.21,0.49,0.74}
\usepackage[pagebackref,breaklinks,colorlinks,allcolors=cvprblue]{hyperref}

\usepackage{amsmath}
\usepackage{multirow}
\usepackage[accsupp]{axessibility}

\def\paperID{*****} 
\def\confName{CVPR}
\def\confYear{2026}

\title{PanoVGGT: Feed-Forward 3D Reconstruction from Panoramic Imagery}

\author{
    Yijing Guo$^1$ \quad Mengjun Chao$^1$ \quad Luo Wang$^1$ \quad Tianyang Zhao$^1$ \\
    Haizhao Dai$^1$ \quad Yingliang Zhang$^2$ \quad Jingyi Yu$^1$ \quad Yujiao Shi$^{1\dagger}$ \\
    $^1$ShanghaiTech University \quad $^2$Sudo \\
}

\begin{document}

\twocolumn[{%
  \renewcommand\tabcolsep{1pt}
  {
   \maketitle
   \vspace{-8mm}
   \begin{center}
     \includegraphics[width=1\textwidth]{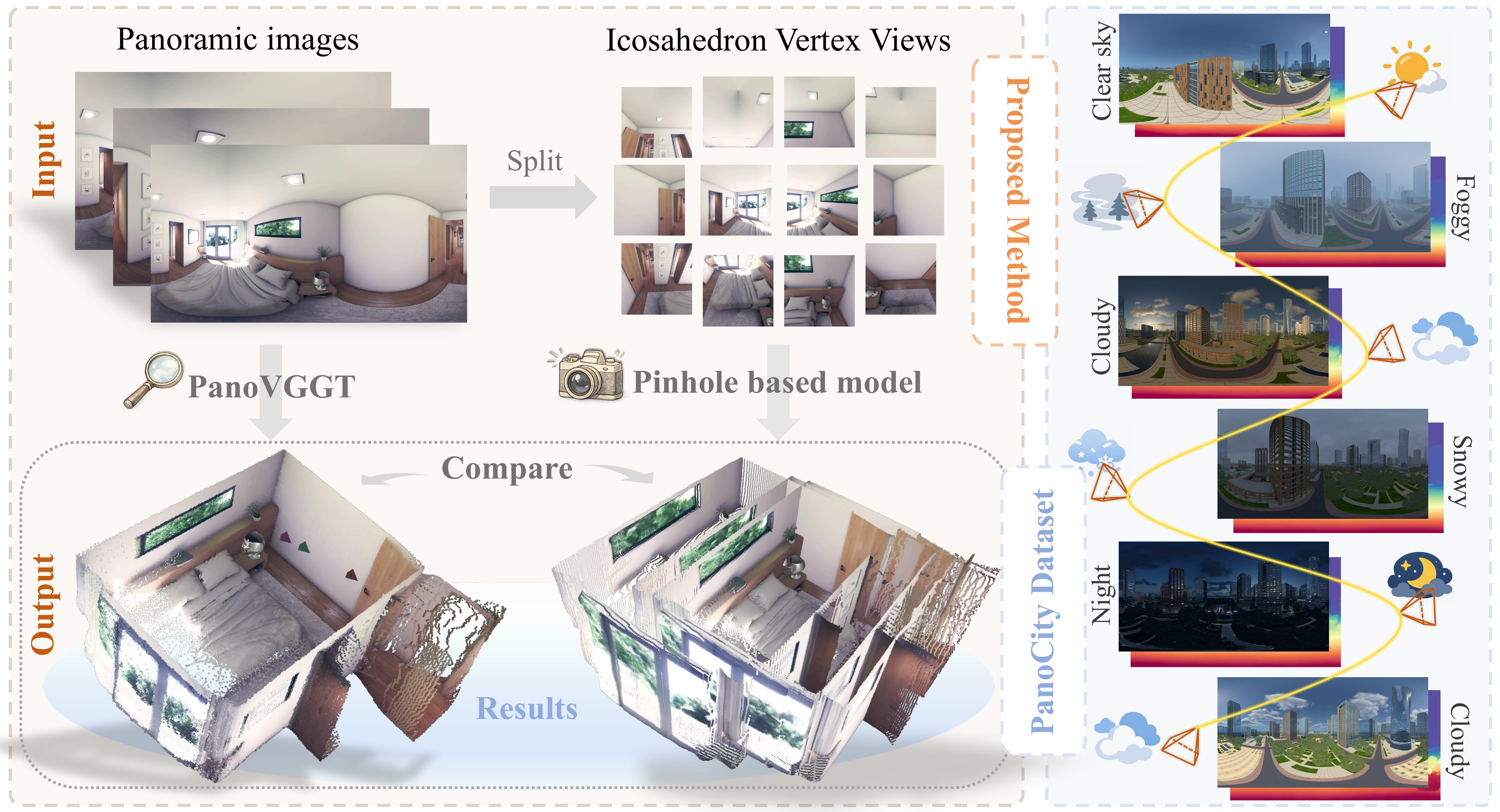}
     \vspace{-6mm}
     \captionof{figure}{
Overview of PanoVGGT and the PanoCity dataset. PanoVGGT predicts camera poses, dense depth, and consistent 3D point clouds from unordered panoramas. Compared with perspective-based pipelines that split panoramas into pinhole views, it yields more accurate 3D reconstructions. PanoCity contains over 120k outdoor panoramas under diverse urban scenes and weather conditions.
     }
     \label{fig:openfigure}
     \vspace{0.2cm}
   \end{center}}%
}]

\begin{abstract}

Panoramic imagery offers a full $360^\circ$ field of view and is increasingly common in consumer devices. However, it introduces non-pinhole distortions that challenge joint pose estimation and 3D reconstruction. Existing feed-forward models, built for perspective cameras, generalize poorly to this setting.
We propose PanoVGGT, a permutation-equivariant Transformer framework that jointly predicts camera poses, depth maps, and 3D point clouds from one or multiple panoramas in a single forward pass. 
The model incorporates spherical-aware positional embeddings and a panorama-specific three-axis 
\(SO(3)\) rotation augmentation, enabling effective geometric reasoning in the spherical domain. To resolve inherent global-frame ambiguity, we further introduce a stochastic anchoring strategy during training. In addition, we contribute PanoCity, a large-scale outdoor panoramic dataset with dense depth and 6-DoF pose annotations. Extensive experiments on PanoCity and standard benchmarks demonstrate that PanoVGGT achieves competitive accuracy, strong robustness, and improved cross-domain generalization. Code and dataset will be released.

\end{abstract}

\section{Introduction}
\label{sec:intro}

Human vision, like most conventional cameras, adheres to a perspective projection model with a limited field of view. In contrast, many animals have near-panoramic eyes that minimize blind spots and enable continuous situational awareness for navigation and threat avoidance. Panoramic coverage offers a geometric advantage for environment understanding: a single vantage can capture the full surround, simplifying global mapping, reducing the need for viewpoint changes, and maintaining consistent exposure and geometry across the scene.

Despite these benefits, most existing vision algorithms remain tailored to perspective imagery. 
Classic solutions, such as stereo matching, structure-from-motion, multi-view stereo, and SLAM systems, rely on multi-stage optimization to recover camera poses and depth. Recent feed-forward approaches, such as DUSt3R~\cite{wang2024dust3r}, VGGT~\cite{wang2025vggt}, and their successors, reformulate this pipeline into an end-to-end paradigm: by leveraging transformers to learn dense cross-view correspondences and geometry priors, they jointly infer depth, pose, and 3D structure in a single forward pass, often surpassing traditional methods in textureless or wide-baseline scenarios. 

However, these models are fundamentally constrained by their perspective assumptions. Recovering a full $360^\circ$ environment requires tiling and stitching multiple crops, introducing seams, parallax inconsistencies, and geometric drift—particularly in dynamic or cluttered scenes. This limitation has spurred the development of panoramic vision methods~\cite{wang2022bifuse++,jiang2021unifuse,sun2021hohonet,yun2023egformer,ai2024elite360d,shen2022panoformer,li2022omnifusion} and datasets~\cite{chang2017matterport3d,armeni2017joint,zheng2020structured3d,liu2022large,albanis2021pano3d,huang2022360vo,huang2024360loc} that adapt neural architectures to spherical imagery. Yet, most of these efforts address isolated tasks (e.g., depth estimation or pose regression) and are limited to small, indoor datasets. End-to-end panoramic 3D reconstruction that unifies multi-view geometry in a single pass remains largely unexplored.

To this end, we introduce PanoVGGT, a feed-forward framework for multi-view 3D reconstruction from unordered panoramic inputs. It jointly predicts camera poses, dense depth maps, and globally consistent 3D point clouds in a single forward pass. By extending permutation-equivariant transformer architectures to the spherical domain through spherical-aware positional embeddings and a panorama-specific three-axis \(SO(3)\) rotation augmentation, PanoVGGT learns to disentangle projection distortions from semantic content and rotation-aware geometric representations without resorting to specialized spherical convolutions.
To resolve the inherent global-frame ambiguity, we adopt a stochastic anchoring strategy that randomly defines the coordinate origin during training, ensuring consistent reconstruction under arbitrary input permutations.

Complementing our method, we construct a new large-scale panoramic dataset, PanoCity, as shown in Fig.~\ref{fig:openfigure}, which contains over 120,000 high-fidelity, high-overlap panoramas with complete pose and depth supervision. The dataset substantially advances the scale and diversity available for panoramic 3D reconstruction research, particularly in outdoor urban environments.

Extensive experiments demonstrate that PanoVGGT achieves strong overall performance across public benchmarks and our new dataset, outperforming prior methods on most metrics while maintaining robust cross-domain generalization. Together, our model and data advance panoramic 3D perception toward unified, efficient, and scalable geometric understanding.

\section{Related Work}
\label{sec:related_work}

\textbf{Feed-forward 3D Reconstruction.}  
Classical SfM and MVS pipelines~\cite{schonberger2016structure,shen2025evolving,wang2025look} deliver accurate reconstruction but rely on iterative optimization and are sensitive to texture and scene dynamics. Recent feed-forward approaches replace these components with end-to-end geometric reasoning. DUSt3R~\cite{wang2024dust3r} directly regresses correspondences and 3D structure from image pairs, and its successors such as MASt3R~\cite{leroy2024grounding} and related variants~\cite{lu2025align3r,zhang2024monst3r,chen2025easi3r} improve robustness and scalability. Large Transformer-based models like VGGT~\cite{wang2025vggt} further extend this paradigm to multi-view input, predicting depth and pose jointly in a single forward pass, while $\pi^3$
~\cite{wang2025pi3} introduces permutation equivariance for order-invariant multi-view inference. Although highly successful on perspective datasets~\cite{li2018megadepth,reizenstein2021common}, these architectures are fundamentally designed under pinhole projection assumptions, limiting their direct applicability to panoramic imagery with spherical distortions.

\vspace{0.4em}
\textbf{Panoramic 3D understanding.}  
Panoramic images capture \(360^\circ\) context but introduce significant geometric distortions and uneven sampling inherent to spherical geometry. 
Prior work addresses these challenges by designing distortion-aware representations, including cubemap or polyhedral re-projection~\cite{wang2020bifuse,wang2022bifuse++,jiang2021unifuse,ai2024elite360d,cao2025panda}, tangent-plane approximations~\cite{shen2022panoformer}, or spherical convolutional operators~\cite{cohen2018spherical}. Most panoramic research focuses on single-task learning: depth estimation via reprojection consistency or distillation~\cite{wang2022bifuse++,wang2024depth,cao2025panda}, and pose estimation through supervised or self-supervised relative pose regression~\cite{tu2024panopose,zheng2025scene}. However, unified feed-forward models that jointly infer pose, depth, and coherent 3D structure from panoramas remain largely unexplored.

\section{Dataset - PanoCity }
\label{dataset}
 
\hypersetup{
    colorlinks=true, 
    urlcolor=magenta,   
}

\begin{table*}[h!]
\vspace{-4mm}
\centering
\setlength{\tabcolsep}{10pt}
\scriptsize
\caption{
Comparison of PanoCity with prior panoramic datasets (\checkmark: available; \texttimes: unavailable). Unlike existing datasets that are limited in scale, annotation completeness, or multi-view continuity, our dataset provides large-scale, high-fidelity, trajectory-based multi-view panoramas with complete 6DoF pose and 16-bit depth supervision.}
\vspace{-2mm}
\small
\begin{tabular}{llllll}
\toprule
Dataset & \#Images & Scene & Pose & Depth & Limitations / Characteristics \\
\midrule
PanoSUNCG \cite{wang2018self} & $\sim$25K & Indoor & {\checkmark~(6DoF)} & \checkmark~(8-bit) & Deprecated and no longer legally available \\
Matterport3D \cite{chang2017matterport3d} & $\sim$10.8K & Indoor & {\checkmark~(6DoF)} & \checkmark~(16-bit) & Discrete panoramas with limited overlap \\
Structured3D \cite{zheng2020structured3d} & $\sim$12K & Indoor & {\checkmark~(3DoF)} & \checkmark~(16-bit) & Lack rotation annotations and no image overlap \\
Stanford2D3D \cite{armeni2017joint} & $\sim$1.4K & Indoor & {\checkmark~(6DoF)} & \checkmark~(16-bit) & Small-scale dataset with limited overlap \\
Pano3D \cite{albanis2021pano3d} & $\sim$42K & Mixed & \texttimes & \checkmark~(16-bit) & No camera pose annotations \\
IndoorLoc \cite{liu2022large} & $\sim$0.7K & Indoor & {\checkmark~(6DoF)} & \checkmark~(16-bit) & RGB–depth inconsistencies and small scale \\
360Loc \cite{huang2024360loc} & $\sim$2.9K & Indoor & {\checkmark~(6DoF)} & \checkmark~(16-bit) & RGB–depth inconsistencies \\
Mapillary Metropolis\textsuperscript{\ref {web}} & $\sim$27K & Outdoor & {\checkmark~(6DoF)} & \checkmark~(sparse) & Sparse and inconsistent depth maps \\
360VO \cite{huang2022360vo} & $\sim$22.5K & Outdoor & {\checkmark~(6DoF)} & \texttimes & Lacks depth annotations \\
\midrule
\textbf{PanoCity (Ours)} & $\sim$120K & Outdoor & {\checkmark~(6DoF)} & \checkmark~(16-bit) & Large scale, high-overlap, and full annotations \\
\bottomrule
\end{tabular}
\label{tab:panodatasets}
\vspace{-4mm}
\end{table*}

Recent feed-forward multi-view reconstruction models for perspective imagery—exemplified by VGGT and $\pi^{3}$—have demonstrated remarkable success by leveraging large-scale, fully annotated, and diverse datasets. However, extending these paradigms to panoramic ($360^\circ$ equirectangular) imagery reveals a fundamental bottleneck: the scarcity of suitable data. Existing panoramic datasets suffer from one or more critical limitations, including insufficient scale, incomplete geometric annotations, RGB–depth inconsistencies, and, crucially, sparse and discrete viewpoint sampling. The lack of stable viewpoint overlap across frames impedes the feed-forward joint learning of dense depth and 6DoF camera motion, which requires not only accurate per-frame depth and pose but also sufficient multi-view overlap to enforce geometric consistency.

~\Cref{tab:panodatasets} compares commonly used panoramic datasets. 
Among these, PanoSUNCG \cite{wang2018self} ($\sim$25K) is no longer legally available for use. Matterport3D \cite{chang2017matterport3d} and Stanford2D3D \cite{armeni2017joint} offer high-quality annotations but primarily consist of discrete panoramas with insufficient inter-frame overlap to support multi-view feed-forward training.
Structured3D \cite{zheng2020structured3d} ($\sim$12K) captures a single panorama with a fixed rotation angle per room, resulting in no overlap between different images and less diversity in pose variations.
IndoorLoc, 360Loc \cite{liu2022large,huang2024360loc}, and Mapillary Metropolis\footnote{\url{https://www.mapillary.com/dataset/metropolis}\label{web}}  suffer from RGB–depth content mismatches.
Pano3D \cite{albanis2021pano3d} lacks camera pose annotations, while 360VO \cite{huang2022360vo} does not include depth data.
Collectively, these shortcomings highlight the urgent need for a comprehensive panorama dataset that simultaneously provides large-scale, complete geometric annotations and continuous multi-view trajectories compatible with feed-forward reconstruction paradigms.

In response to this need, we introduce PanoCity, a large-scale, photorealistic panoramic dataset focused on ground-level urban scenes. Data is fully collected in Unreal Engine 5 (UE5)~\cite{engine2018unreal} integrated with AirSim~\cite{shah2017airsim}, allowing precise, controllable camera motion. Cameras follow smooth trajectories along realistic city roads, constrained by vehicle and pedestrian kinematics to ensure stable inter-frame translation, rotation, and sufficient viewpoint overlap. To minimize domain bias, we select three diverse virtual cities featuring varied urban textures and architectural styles. Beyond scene variety, PanoCity is rendered under five diverse weather and illumination conditions: \emph{sunny noon, partly cloudy dusk, foggy morning, evening,} and \emph{snowy}. This extensive variation compels models to learn robust geometric representations rather than relying on low-level texture or lighting biases.

PanoCity contains over 120,000 frames. Each frame includes a high-resolution $4096 \times 2048$ equirectangular RGB image, a metrically consistent 16-bit depth map, and an accurate 6DoF camera-to-world pose. RGB, depth, and pose are rendered synchronously and are therefore perfectly aligned, enabling reliable supervision for joint learning of depth, pose, and 3D reconstruction.

Overall, PanoCity substantially expands the scale and diversity of synthetic panoramic datasets. By combining accurate geometric annotations, continuous multi-view trajectories, and diverse urban conditions, it provides a stronger foundation for learning pose, depth, and 3D reconstruction from panoramic imagery.

\section{Method – PanoVGGT}
\label{method}

\begin{figure*}[t]
    \centering
    \includegraphics[width=1.0\textwidth]{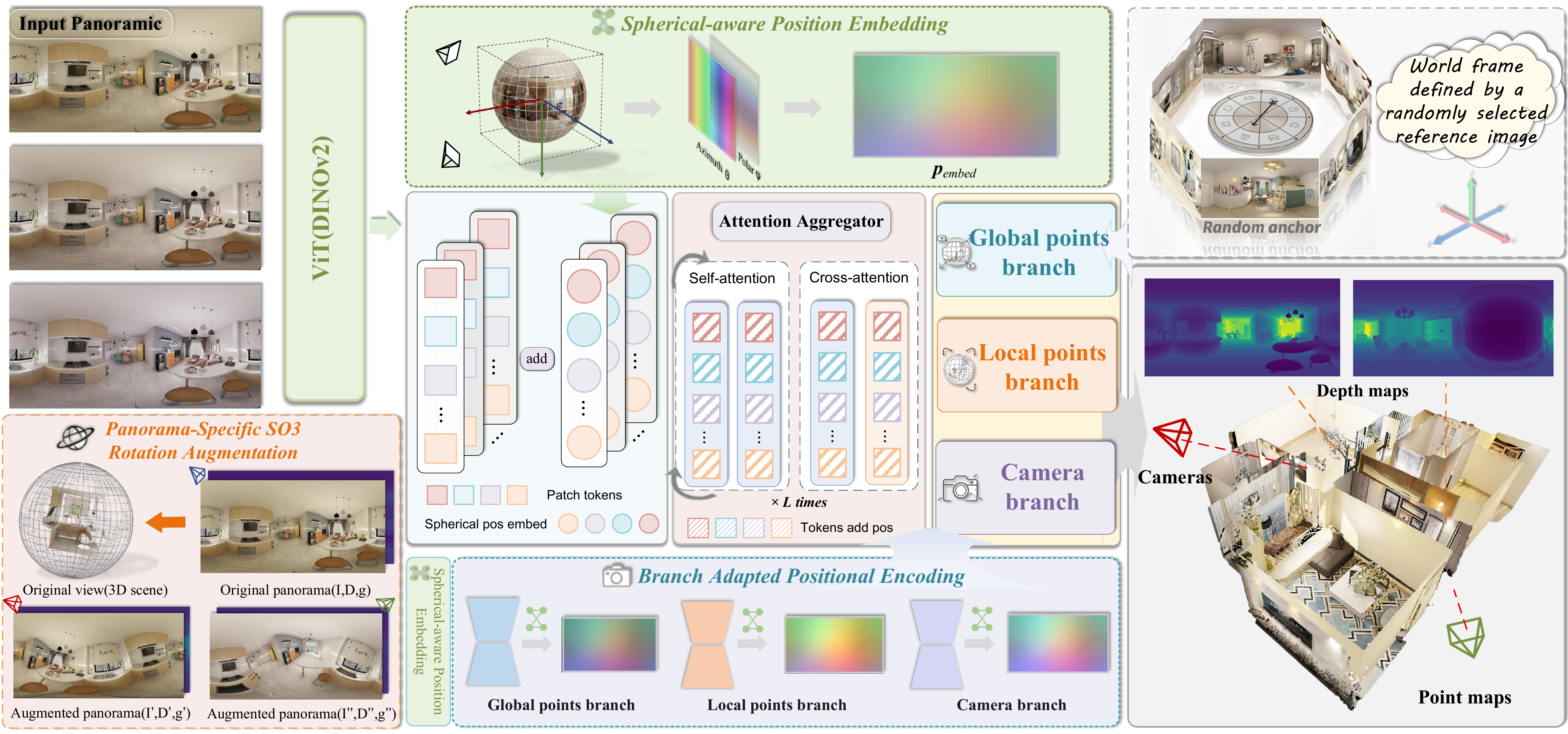}
    \vspace{-7mm}
    \caption{Overview of the proposed PanoVGGT framework. Given one or multiple panoramic images, PanoVGGT extracts spherical patch tokens using a ViT backbone, augments them via $SO(3)$ geometric augmentation, and encodes them with shared and branch-adapted positional embeddings. A multi-branch attention aggregator jointly reasons about local geometry, global structure, and camera motion. The model outputs dense depth maps, camera poses (under a randomly selected global anchor), and local/global point maps for final 3D reconstruction.}
    \label{fig:arch}
    \vspace{-4mm}
\end{figure*}

In this section, we present PanoVGGT, a feed-forward 3D reconstruction framework tailored to panoramic imagery. Given an unordered set of panoramas, PanoVGGT jointly predicts camera poses, dense depth maps, and globally consistent 3D point clouds in a single forward pass. By extending the large-scale Transformer feed-forward paradigm---previously validated on perspective images by VGGT and $\pi^3$---we adapt its core principles to the fundamentally different geometry of full-spherical panoramas.

\subsection{Framework Overview}
\label{sec:arch}

PanoVGGT is designed to learn a mapping
\begin{equation}
f_\theta: I = \{I_i\}_{i=1}^N \rightarrow (G, D, P),
\end{equation}
where the input is an \emph{unordered} collection of panoramas \(I_i \in \mathbb{R}^{H \times W \times 3}\). The model jointly predicts the camera poses, dense depth maps, and world-frame 3D point clouds for each input view. Specifically, camera poses are represented as \(G = \{g_i\}_{i=1}^N\), with each \(g_i \in SE(3)\) encoding the camera-to-world transformation. Depth maps \(D = \{D_i\}_{i=1}^N\) provide per-pixel Euclidean distances, while the 3D point clouds \(P = \{P_i\}_{i=1}^N\), where \(P_i \in \mathbb{R}^{H \times W \times 3}\), specify 3D locations expressed in a shared, globally consistent coordinate frame.

Although the 3D points can be computed directly from the predicted poses and depths, PanoVGGT explicitly regresses them as an additional output. This redundancy, termed over-complete supervision, has been shown to substantially enhance the accuracy and robustness of all predictions, echoing similar findings in VGGT and \(\pi^3\).

Since the inputs form a set rather than a sequence, their ordering must not influence the reconstruction. PanoVGGT is therefore constructed to be {permutation-equivariant}: reordering $\{I_i\}$ produces a correspondingly reordered set of predicted poses, depths, and point clouds. This property eliminates the need for a designated reference frame.

The overall model design, illustrated in ~\cref{fig:arch}, follows an encode--aggregate--decode structure. Each panorama is first passed through a DINOv2 encoder~\cite{oquab2023dinov2} that converts image patches into tokens. These tokens are then added with \emph{Spherical-aware Position Embedding} and processed by a \emph{Geometric Aggregator} composed of a Transformer with alternating intra-view and global attention, enabling joint reasoning over both local geometric cues and cross-view relationships. Finally, multi-task prediction heads decode the aggregated representation into camera poses, per-view depth, and both local and global point clouds, completing the reconstruction pipeline.

\subsection{Spherical-aware Position Embedding}

Transformers require positional encoding, yet standard ViT embeddings are geometrically incompatible with equirectangular panoramas due to their spatially varying sampling density. We therefore introduce a {spherical-aware positional embedding}.

For each patch, we compute its center coordinates $(\theta, \phi)$ and encode them as a circularly symmetric 4D vector:
\begin{equation}
    p_{\mathrm{vec}} = 
\bigl[
\sin\theta,\; \cos\theta,\; \sin\phi,\; \cos\phi
\bigr].
\end{equation}
A lightweight MLP then produces the final embedding
\begin{equation}
    p_{\mathrm{embed}} = \mathrm{MLP}(p_{\mathrm{vec}}) \in \mathbb{R}^C.
    \label{eq:spe}
\end{equation}

This encoding inherently preserves wrap-around continuity at the angular boundaries $\theta = \pm \pi$, ensuring smooth transitions across the panorama seam. This property is essential for maintaining consistent spatial relationships in equirectangular projections, where naive embeddings often fail due to discontinuities at the image edges.

More importantly, this design synergizes deeply with our proposed Geometric $SO(3)$ augmentation strategy described in ~\cref{sec:aug}. In this augmentation, the spatial grid positions remain fixed, while the visual content rotates freely over the spherical domain. Consequently, patches located at a fixed grid position (e.g., near the top of the sphere, where $\theta \approx -\pi/2$) receive a constant positional embedding, yet the actual scene content within those patches may originate from any direction in the original panorama, including regions originally at the sides or bottom.

This deliberate decoupling between fixed position embeddings and dynamically rotated content forces the model to \emph{disentangle the effects of projection distortion from the underlying scene semantics}. Instead of relying on heuristic biases---such as assuming that the top grid patches always correspond to the sky or ceiling---the network learns to treat the positional embedding as a modulating signal that informs how to interpret visual features under specific local distortion patterns. Through continuous spherical rotations, the model learns to infer the appropriate distortion characteristics relative to its current orientation, enabling it to align locally observed visual patterns with globally consistent camera poses. Consequently, the network acquires robust, $SO(3)$-equivariant geometric representations without requiring complex architectural modifications such as spherical convolutions.


\subsection{Geometry Aggregator}
\label{sec:aggregator}

For the geometry aggregator, we use $L$ Alternating-Attention blocks, following the design principles of $\pi^3$ and VGGT. Each block consists of two stages. The first stage applies frame-wise self-attention, where tokens attend only to others originating from the same panorama, allowing the network to capture intra-view structures as well as distortions introduced by the equirectangular projection. The second stage performs global self-attention over tokens from all panoramas, enabling cross-view correspondence reasoning and promoting geometric consistency across the entire set. The aggregator does not rely on any positional encoding or cue that specifies view ordering, ensuring that the entire module remains strictly permutation-equivariant. 

After Alternating-Attention, we combine these aggregated features with the learned spherical embeddings from Eq.~\ref{eq:spe}. The spherical embeddings are first transformed by three lightweight adaptors and then concatenated with the aggregated features, which are subsequently fed into three parallel prediction heads. 
The first regresses the camera poses, producing $SE(3)$ transformations $\{g_i\}$. The second predicts local point clouds in the camera coordinate system by applying the spherical ray model to the estimated depth maps $D_i$. The third directly outputs world-frame point clouds $P_i^{\text{global}}$, providing an additional geometric constraint that complements the pose and depth predictions. 

\vspace{0.4em}
\noindent\textbf{Stochastic anchoring.} A key challenge arises from the permutation-equivariant design: the global coordinate frame is fundamentally ambiguous because no canonical reference view exists. Fixing the first frame as the origin, as done in VGGT, introduces an ordering bias that can destabilize pose prediction in unordered input scenarios. To enhance pose estimation stability, we employ a \emph{stochastic anchoring} strategy. In each training iteration, a random panorama $k$ is selected as the anchor, and all other poses and point clouds are consistently aligned to this anchor-centered coordinate frame. This randomized re-centering enforces a stable ``center–spoke'' geometric structure across training steps while preserving full permutation equivariance~\cite{zaheer2017deep} with respect to the input set.

\subsection{Panorama-Specific Data Augmentation}
\label{sec:aug}

Unlike perspective images, panoramas naturally support full spherical rotations, allowing augmentation directly in $SO(3)$ without breaking geometric validity. We exploit this property through a \emph{three-axis rotation augmentation} applied jointly to RGB images, depth maps, and camera poses. For each RGB--Depth--Pose triplet $(I_i, D_i, g_i)$, a random rotation $R_{\mathrm{aug}} \in SO(3)$ is first sampled, and the camera pose is updated to $g_i' = R_{\mathrm{aug}} \cdot g_i$. The original panorama and its depth map are then projected onto the unit sphere, rotated by $R_{\mathrm{aug}}$, and finally re-sampled back into equirectangular form to obtain $(I_i', D_i')$.

This process yields new, physically correct viewpoints for every panorama, effectively enabling unbounded data augmentation. Beyond alleviating data scarcity, spherical rotation augmentation compels the model to acquire true rotational robustness rather than exploiting biases inherent to the equirectangular projection.


\subsection{Loss Functions}
\label{sec:loss}

PanoVGGT is trained end-to-end using a multi-task objective that jointly supervises depth (local geometry), point cloud (global geometry), and camera poses.

\vspace{0.4em}
\noindent\textbf{Scale-consistent local and global geometry loss.}
Following $\pi^3$~\cite{wang2025pi3}, we estimate a single optimal scale $s^\ast$ per training sample to ensure metric consistency across all geometric predictions.
For each local point head, $s^\ast$ is obtained by aligning the predicted camera-frame points $\hat{P}_i^{\text{local}}$ to the ground-truth points $P_i^{\text{local}}$:
\begin{equation}
s^\ast
= \arg\min_{s>0} \sum_i
\bigl\|\, s\, \hat{P}_i^{\text{local}} - P_i^{\text{local}} \bigr\|_2^2 ,
\end{equation}
whose closed-form solution is
\begin{equation}
s^\ast =
\frac{\sum_i \langle \hat{P}_i^{\text{local}},\, P_i^{\text{local}} \rangle}
     {\sum_i \| \hat{P}_i^{\text{local}} \|_2^2},
\qquad
s^\ast \leftarrow \max(|s^\ast|,\, 10^{-6}).
\end{equation}
This scale is shared across all geometric heads, and the scale-aligned local and global geometry loss is defined as

\begin{equation}
\begin{aligned}
\mathcal{L}_{\text{lp}} &=
\sum_i \| s^\ast \hat{P}_i^{\text{local}} - P_i^{\text{local}} \|_1, \\
\mathcal{L}_{\text{gp}} &=
\sum_i \| s^\ast \hat{P}_i^{\text{global}} - P_i^{\text{global}} \|_1.
\end{aligned}
\end{equation}

\noindent\textbf{Normal consistency regularization.}
To encourage locally smooth yet geometrically faithful surfaces, we apply a normal consistency loss $\mathcal{L}_{\text{nor}}$. 
Surface normals are estimated from neighboring 3D points, and their angular differences are penalized using a SmoothL1 function:
\begin{gather}
\mathcal{L}_{\text{nor}} =
\frac{1}{|\Omega|}
\sum_{p \in \Omega}
\operatorname{SmoothL1}\!\left(
\angle(\hat{n}_{p},\, n_{p})
\right),\\[4pt]
\angle(a,b) =
\arccos\!\left(
\tfrac{a^\top b}{\|a\|\|b\|}
\right),
\end{gather}
where $\Omega$ denotes valid pixels excluding depth-edge regions.

\vspace{0.4em}
\noindent\textbf{Relative pose supervision.}
Let $\hat{R}_{i\leftarrow j}$ and $\hat{t}_{i\leftarrow j}$ denote the predicted relative rotation and translation between frames $j$ and $i$, and
$R_{i\leftarrow j}^{k}$ and $t_{i\leftarrow j}^{k}$ their corresponding ground-truth transformations obtained via stochastic anchoring.
Rotations are supervised via the angular distance:
\begin{equation}
\mathcal{L}_{\text{rot}} = \angle(\hat{R}_{i\leftarrow j},\, R^k_{i\leftarrow j}),
\end{equation}
\begin{equation}
\angle(A,B) =
\arccos\!\left(\tfrac{\operatorname{tr}(A^\top B)-1}{2}\right).
\end{equation}
Translations are supervised in the same scale-consistent space:
\begin{equation}
\mathcal{L}_{\text{trans}}
= \| s^\ast \hat{t}_{i\leftarrow j} - t_{i\leftarrow j}^{k} \|_1 .
\end{equation}

\vspace{0.4em}
\noindent\textbf{Overall objective.}
The final training loss combines all components:
\begin{equation}
\mathcal{L} =
\mathcal{L}_{\text{lp}}
+ \mathcal{L}_{\text{gp}}
+ \mathcal{L}_{\text{nor}}
+ \lambda_{\text{g}}\!\left(
\lambda_{\text{T}} \mathcal{L}_{\text{trans}}
+ \mathcal{L}_{\text{rot}}
\right),
\end{equation}
where $\lambda_{\text{T}}{=}100$ and $\lambda_{\text{g}}{=}0.1$.

\begin{table*}[t]
\centering
\caption{Camera pose estimation results. 
\textbf{Bold} = best, \underline{underline} = second best.
}
\label{tab:pose}
\vspace{-3mm}
\resizebox{\textwidth}{!}{%
\begin{tabular}{l c cccc ccccc ccccc}
\toprule
\multirow{2}{*}{Method} &
\multicolumn{5}{c}{\textbf{Matterport3D} (Indoor)} &
\multicolumn{5}{c}{\textbf{Stanford2D3D} (Indoor)} &
\multicolumn{5}{c}{\textbf{PanoCity} (Outdoor)} \\
\cmidrule(lr){2-6} \cmidrule(lr){7-11} \cmidrule(lr){12-16}
& AUC@30$\uparrow$ &
\multicolumn{2}{c}{Rotation ($^\circ\!\downarrow$)} &
\multicolumn{2}{c}{Translation ($^\circ\!\downarrow$)} &
AUC@30$\uparrow$ &
\multicolumn{2}{c}{Rotation ($^\circ\!\downarrow$)} &
\multicolumn{2}{c}{Translation ($^\circ\!\downarrow$)} &
AUC@30$\uparrow$ &
\multicolumn{2}{c}{Rotation ($^\circ\!\downarrow$)} &
\multicolumn{2}{c}{Translation ($^\circ\!\downarrow$)} \\
\cmidrule(lr){3-4} \cmidrule(lr){5-6}
\cmidrule(lr){8-9} \cmidrule(lr){10-11}
\cmidrule(lr){13-14} \cmidrule(lr){15-16}
& & Mean & Med & Mean & Med & & Mean & Med & Mean & Med & & Mean & Med & Mean & Med \\
\midrule
Bifusev2~\cite{wang2022bifuse++}       
& 0.007 & 66.049 & 60.924 & 82.038 & 88.794 
& 0.030 & 57.568 & 43.296 & 45.933 & 45.460 
& \underline{0.833} & 1.655 & \textbf{0.173} & \underline{5.044} & \underline{4.731} \\

VGGT~\cite{wang2025vggt}           
& 0.034 & 66.510 & 56.329 & 47.857 & 48.149 
& 0.047 & 71.035 & 61.112 & 41.540 & 40.011 
& 0.205 & 7.659 & 1.609 & 35.867 & 28.796 \\

$\pi^3$~\cite{wang2025pi3}        
& 0.047 & 63.286 & 54.292 & 44.201 & 42.750 
& 0.076 & 58.727 & 45.378 & 39.567 & 35.652 
& 0.571 & 7.669 & 1.020 & 16.780 & 7.768 \\

$\pi^3$*  
& \underline{0.305} & \underline{25.205} & \underline{25.205} & \underline{31.550} & \underline{31.550} 
& \underline{0.274} & \underline{19.491} & \underline{24.173} & \underline{33.516} & \underline{27.871} 
& 0.682 & \textbf{0.525} & \underline{0.474} & 22.379 & 15.727 \\

\midrule
PanoVGGT (ours)  
& \textbf{0.459} & \textbf{21.394} & \textbf{23.466} & \textbf{18.900} & \textbf{18.100} 
& \textbf{0.556} & \textbf{18.801} & \textbf{24.170} & \textbf{10.999} & \textbf{9.762} 
& \textbf{0.949} & \underline{0.873} & 0.834 & \textbf{2.168} & \textbf{1.684} \\

\bottomrule
\end{tabular}%
}
\end{table*}

\section{Experiments}
\label{sec:results}

We evaluate PanoVGGT on three core tasks—camera pose estimation, depth prediction, and point cloud reconstruction—using a unified experimental protocol. All models, including baselines, are trained and assessed under identical settings to ensure fair comparison. We use five datasets: Matterport3D~\cite{chang2017matterport3d}, Stanford2D3D~\cite{armeni2017joint}, Structured3D~\cite{zheng2020structured3d}, Pano3D~\cite{albanis2021pano3d,xia2019gibson} and PanoCity, spanning a broad spectrum of indoor and outdoor environments. All panoramas are resized to $512{\times}1024$ for both training and evaluation.

\vspace{0.4em}
\noindent\textbf{Training \& testing splits of standard benchmarks.}
The original dataset splits of Matterport3D, Stanford2D3D, and Structured3D are not designed for multi-view supervision with paired depth and relative poses. To obtain valid training samples, we re-partition all datasets.
For Matterport3D and Stanford2D3D, we construct new splits by detecting panorama groups with reliable viewpoint overlap using metadata and subsequent manual verification. Structured3D presents an additional difficulty: each room contains only a single panorama, and true cross-room overlap is extremely rare, resulting in minimal usable multi-view data.
Given the limited number of valid multi-view samples, we adopt an $8{:}1{:}1$ split at the scene level, ensuring strict separation across training, validation, and testing scenes. All baselines are retrained using these revised splits. Despite careful curation, many indoor pairs still exhibit weak or no overlap due to occlusions and complex room layouts, making these datasets inherently challenging for camera pose estimation.

\vspace{0.4em}
\noindent\textbf{Training and testing splits of PanoCity.}
PanoCity consists of kilometer-scale trajectories, each containing hundreds to thousands of panoramas. We split the dataset at the trajectory level to prevent any image leakage across splits, resulting in an approximate $18{:}1{:}1$ distribution for training, validation, and testing.
This trajectory-level split maximizes data usage while preserving strong domain shift: held-out trajectories span diverse urban morphologies, including dense metropolitan centers, residential areas, campuses, parks, and highway corridors. This structure provides a stringent benchmark for evaluating generalization in large-scale outdoor environments.

\vspace{0.4em}
\noindent\textbf{Implementation Details.}
Training is performed on 8 NVIDIA~A100 GPUs for roughly ten days, using input panoramas at a resolution of $336\times672$. 
During training, we dynamically sample sequences of length $N\in[2,24]$, and each GPU processes up to 384 images with eight-step gradient accumulation to maintain memory efficiency at scale.

\begin{table*}[t]
\centering
\scriptsize
\setlength{\tabcolsep}{8pt}
\caption{Monocular Depth estimation performance. 
\textbf{Bold} = best, \underline{underline} = second best.}
\label{tab:depth_in_domain}
\vspace{-3mm}
\begin{tabular}{llcccccccccc}
\toprule
\textbf{Model} & \textbf{Input} &
\multicolumn{2}{c}{\textbf{Matterport3D} (Indoor)} &
\multicolumn{2}{c}{\textbf{Stanford2D3D} (Indoor)} &
\multicolumn{2}{c}{\textbf{Structured3D} (Indoor)} &
\multicolumn{2}{c}{\textbf{PanoCity} (Outdoor)} \\
\cmidrule(lr){3-4} \cmidrule(lr){5-6} \cmidrule(lr){7-8} \cmidrule(lr){9-10}
& & Abs Rel $\downarrow$ & $\delta < 1.25$ $\uparrow$
& Abs Rel $\downarrow$ & $\delta < 1.25$ $\uparrow$
& Abs Rel $\downarrow$ & $\delta < 1.25$ $\uparrow$
& Abs Rel $\downarrow$ & $\delta < 1.25$ $\uparrow$ \\
\midrule
UniFuse~\cite{jiang2021unifuse}      & Monocular & 0.1367 & 0.8314 & 0.1390 & 0.8270 & 0.0387 & 0.9804 & 1.0374 & 0.4365  \\
BiFuse++~\cite{wang2022bifuse++}     & Monocular & 0.1076 & 0.8846 & 0.1120 & 0.8805 & \underline{0.0293} & 0.9840 & \underline{0.0200} & \underline{0.9843} \\
HoHoNet~\cite{sun2021hohonet}      & Monocular & 0.1217 & 0.8576 & 0.1192 & 0.8872 & 0.0401 & 0.9759 & 0.0297 & \textbf{0.9885} \\
EGFormer~\cite{yun2023egformer}     & Monocular & 0.0987 & 0.9082 & 0.0929 & 0.8890 & \textbf{0.0273} & \textbf{0.9889} & 0.0363 & 0.9807 \\
Elite360D~\cite{ai2024elite360d}    & Monocular & 0.1263 & 0.8534 & 0.1273 & 0.8642 & 0.1019 & 0.9180 & 0.3622 & 0.4171 \\
PanoFormer~\cite{shen2022panoformer}   & Monocular & 0.1007 & 0.8985 & 0.1060 & 0.8887 & 0.0328 & 0.9823 & 0.0765 & 0.9680 \\
OmniFusion~\cite{li2022omnifusion}   & Monocular & 0.1614 & 0.8086 & 0.1420 & 0.8261 & 0.0471 & 0.9754 & 0.0362 & 0.9779 \\
\midrule
PanoVGGT (Ours) & Monocular    
& \underline{0.0884} & \underline{0.9157} 
& \textbf{0.0711} & \textbf{0.9392} 
& 0.0438 & 0.9728 
& 0.0312 & 0.9713  \\

PanoVGGT (Ours) & Multi-view   
& \textbf{0.0840} & \textbf{0.9266} 
& \underline{0.0778} & \underline{0.9323} 
& 0.0400 & \underline{0.9870} 
& \textbf{0.0196} & 0.9812  \\
\bottomrule
\end{tabular}
\vspace{-4mm}
\end{table*}

\begin{figure*}[t]
\centering
\includegraphics[width=\textwidth]{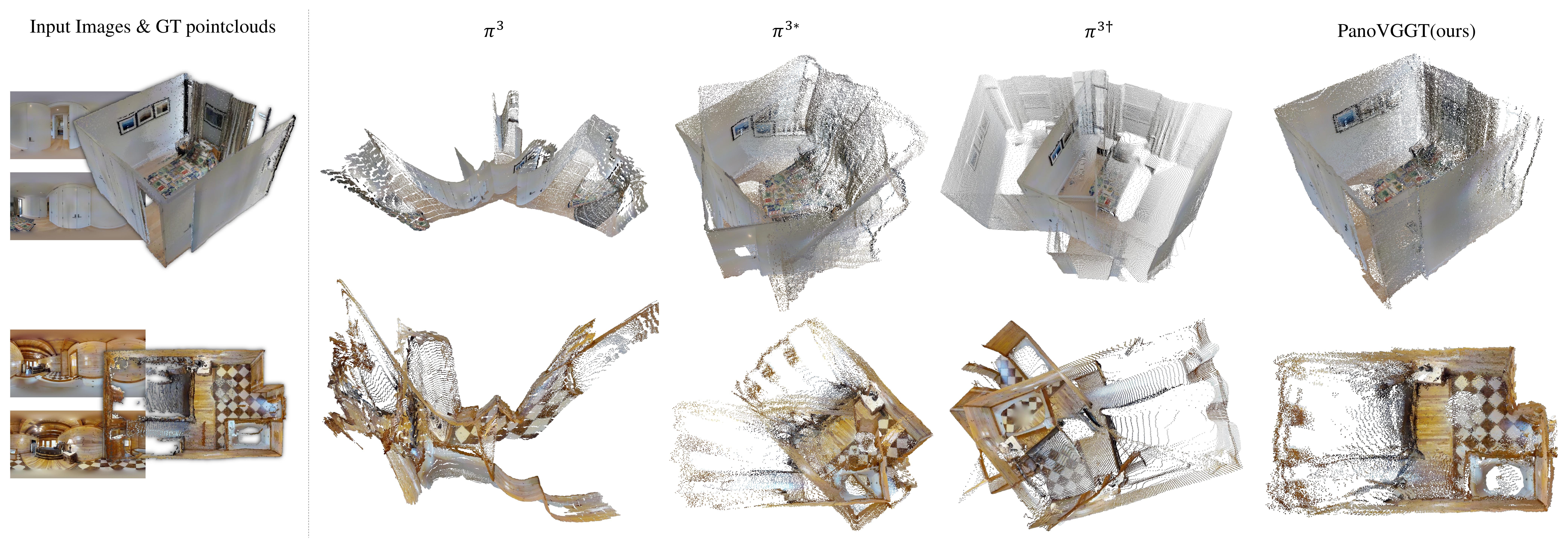}
\vspace{-4mm}
\caption{Multi-view point-cloud reconstructions on the Matterport3D dataset from two unordered panoramic inputs.
From left to right: $\pi^3$, $\pi^3$*, $\pi^3{}^\dagger$, and PanoVGGT.
Here, $\dagger$ denotes the original pinhole-only $\pi^3$ applied to panoramic inputs via MoGe's dodecahedral projection protocol, which decomposes each panorama into 12 perspective views.
PanoVGGT produces sharper and more structurally consistent indoor reconstructions.}
\label{fig:supp_vis_matterport}
\end{figure*}

\subsection{Camera Pose Estimation}

Camera pose estimation is evaluated on Matterport3D, Stanford2D3D, and PanoCity. Structured3D is excluded because it lacks multi-view panoramas with sufficient overlap for reliable relative pose estimation. We compare against Bifusev2~\cite{wang2022bifuse++}, VGGT~\cite{wang2025vggt}, and $\pi^3$~\cite{wang2025pi3}. Bifusev2 is evaluated using the official implementation. VGGT and $\pi^3$ are tested directly on panoramic images using their released weights. In addition, we re-train $\pi^3$ on panoramic data using the same protocol as our method to assess its adaptability to $360^\circ$ imagery; this variant is denoted as $\pi^3$*.

For fair comparison, all multi-view experiments follow a three-view protocol. This choice is dictated by dataset constraints: most scenes in Matterport3D and Stanford2D3D contain only three panoramas. Using more views would drastically reduce the number of valid test samples. Moreover, even with three views, mutual overlap is not guaranteed, making pose estimation particularly challenging.

We report the area under the accuracy curve (AUC) at a $30^\circ$ threshold, together with mean and median rotation and translation errors. Both rotation and translation are evaluated using angular errors (in degrees), avoiding scale ambiguity and enabling consistent comparison across datasets. All metrics are computed after filtering zero-translation pairs to remove degenerate cases.

As shown in \cref{tab:pose}, PanoVGGT achieves the highest AUC and the lowest rotation and translation errors in most settings, demonstrating strong geometric consistency and robust pose estimation. Despite being designed for panoramic imagery, Bifusev2 performs worse than VGGT and $\pi^3$ on Matterport3D and Stanford2D3D. This is largely because its self-supervised training relies on ordered frames with narrow baselines and substantial overlap. In contrast, panoramas in these datasets are captured from sparse and unordered viewpoints, often with limited or no overlap, violating these assumptions and leading to unstable pose estimation. VGGT, on the other hand, benefits from large-scale pretraining on diverse image data, and pixels near the equator of panoramas locally approximate pinhole geometry, allowing it to generalize more effectively to panoramic inputs. This observation highlights the importance of training strategies tailored for sparse panoramic viewpoints.

\subsection{Monocular Depth Estimation}

We evaluate depth on five datasets against baselines including UniFuse~\cite{jiang2021unifuse}, BiFuse++~\cite{wang2022bifuse++}, HoHoNet~\cite{sun2021hohonet}, EGFormer~\cite{yun2023egformer}, Elite360D~\cite{ai2024elite360d}, PanoFormer~\cite{shen2022panoformer}, and OmniFusion~\cite{li2022omnifusion}. Predictions are scale-normalized using IRLS. As shown in \cref{tab:depth_in_domain}, our monocular setting achieves competitive accuracy, which is further improved using multi-view inputs. Notably, while BiFuse++ and EGFormer show slight advantages on Structured3D and PanoCity, they rely on separate, depth-only models trained per dataset. In contrast, PanoVGGT is a single, unified model performing joint pose-depth-point estimation across all datasets simultaneously. Despite this highly generalized multi-task setting, our model achieves state-of-the-art performance on Matterport3D and Stanford2D3D.

\begin{table}[t]
\centering
\setlength{\tabcolsep}{9pt}
\scriptsize
\caption{Zero-shot depth estimation on \textbf{Pano3D (GibsonV2)} under scale-only (top) and scale+shift (bottom) alignments. 
$\dagger$ denotes pinhole-based models evaluated on panoramas using the dodecahedral projection protocol of MoGe.}
\label{tab:zero_shot_pano3d_compact}
\vspace{-2mm}
\begin{tabular}{lcccc}
\toprule
Method & Abs Rel ↓ & RMSE ↓ & $\delta_1$ ↑ & $\delta_2$ ↑ \\
\midrule
VGGT$^\dagger$~\cite{wang2025vggt}     & 0.1266 & 0.3402 & 0.8650 & 0.9816 \\
$\pi^3$$^\dagger$  & \underline{0.1196} & 0.3145 & \underline{0.8811} & \underline{0.9855} \\
MoGe$^\dagger$~\cite{wang2025moge}     & 0.1236 & 0.3353 & 0.8589 & 0.9797 \\
MoGeV2$^\dagger$~\cite{wang2025moge2}   & 0.1264 & 0.3429 & 0.8578 & 0.9799 \\
PanDA~\cite{cao2025panda}              & 0.3577 & 0.7506 & 0.4069 & 0.6245 \\
Unik3D~\cite{piccinelli2025unik3d}             & 0.1568 & \underline{0.3107} & 0.7989 & 0.9636 \\
\textbf{PanoVGGT}  & \textbf{0.0869} & \textbf{0.3069} & \textbf{0.9223} & \textbf{0.9859} \\
\midrule
VGGT$^\dagger$~\cite{wang2025vggt}     & 0.1224 & 0.3196 & 0.8780 & 0.9815 \\
$\pi^3$$^\dagger$~\cite{wang2025pi3}  & 0.1066 & \underline{0.2972} & 0.9055 & \underline{0.9890} \\
MoGe$^\dagger$~\cite{wang2025moge}     & 0.1279 & 0.3209 & 0.8654 & 0.9792 \\
MoGeV2$^\dagger$~\cite{wang2025moge2}   & 0.1299 & 0.3296 & 0.8581 & 0.9783 \\
PanDA~\cite{cao2025panda}              & \underline{0.0839} & 0.3651 & \underline{0.9241} & 0.9796 \\
Unik3D~\cite{piccinelli2025unik3d}             & 0.1316 & \textbf{0.2616} & 0.8279 & 0.9822 \\
\textbf{PanoVGGT}  & \textbf{0.0833} & 0.3015 & \textbf{0.9299} & \textbf{0.9894} \\
\bottomrule
\end{tabular}
\vspace{-3mm}
\end{table}

To assess cross-dataset generalization, we conduct zero-shot evaluation on Pano3D (GibsonV2), with results reported in Tab.~\ref{tab:zero_shot_pano3d_compact}. The first four methods in the table are designed for pinhole imagery, while the remaining ones are native panoramic models. Following the MoGe protocol, pinhole-based methods are applied to panoramas via dodecahedral projection, which decomposes the image into 12 perspective views with lower distortion than cube-map projections. We report results under both scale-only and scale+shift alignment. PanoVGGT consistently outperforms prior methods across most metrics, demonstrating strong robustness under distribution shift.

\subsection{Point Cloud Reconstruction}
No prior methods are specifically designed for point cloud reconstruction from panoramic imagery. Therefore, we compare PanoVGGT against $\pi^3$, its re-trained panoramic variant $\pi^3$*, and a pinhole-split variant $\pi^3$$^\dagger$. Due to space constraints, we report results on the PanoCity dataset using 10 input panoramas, with additional evaluations on other datasets provided in the supplementary material. We adopt standard point cloud metrics, including mean and median distances for accuracy, completion, and overall quality. As shown in \cref{tab:pointcloud_results}, PanoVGGT achieves consistently lower errors across all metrics, demonstrating more accurate geometry reconstruction and improved scene completeness.

\begin{table}[t]
\centering

\caption{Point cloud reconstruction performance on {PanoCity}. $\dagger$ denotes the perspective-based model evaluated using the dodecahedral projection protocol of MoGe.}
\label{tab:pointcloud_results}
\vspace{-2mm}
\resizebox{\columnwidth}{!}{
\begin{tabular}{lcccccc}
\toprule
\multirow{2}{*}{Method} &
\multicolumn{2}{c}{Acc $\downarrow$} &
\multicolumn{2}{c}{Comp $\downarrow$} &
\multicolumn{2}{c}{Overall $\downarrow$} \\
\cmidrule(lr){2-3} \cmidrule(lr){4-5} \cmidrule(lr){6-7}
& Mean & Med & Mean & Med & Mean & Med \\
\midrule
$\pi^3$~\cite{wang2025pi3}               & 1.404 & 0.905 & 11.737 & 6.002 & 6.570 & 3.454 \\
$\pi^3$$^\dagger$~\cite{wang2025pi3}     & 1.261 & 0.995 & 15.182 & 9.578 & 8.222 & 5.287 \\
$\pi^3$*              & 1.111 & 0.707 & 1.037 & 0.592 & 1.074 & 0.650 \\
\midrule
PanoVGGT (global points) & \underline{0.768} & \textbf{0.179} & \textbf{0.705} & \textbf{0.149} & \textbf{0.737} & \textbf{0.164} \\
PanoVGGT (local points)  & \textbf{0.744} & \underline{0.180} & \underline{0.756} & \underline{0.204} & \underline{0.750} & \underline{0.192} \\
\bottomrule
\end{tabular}
}
\end{table}

\begin{table}[t]
\centering
\setlength{\tabcolsep}{3pt}
\scriptsize
\caption{Ablation study of PanoVGGT components on Matterport3D.
\textbf{Bold} = best, \underline{underline} = second best.
}
\vspace{-2mm}
\label{tab:ablation}
\begin{tabular}{lcccc}
\toprule
\multirow{2}{*}{Components} &
\multicolumn{2}{c}{Pose} &
\multicolumn{2}{c}{Depth} \\
\cmidrule(lr){2-3} \cmidrule(lr){4-5}
& AUC@30↑ & R$_{\text{mean}}$/T$_{\text{mean}}$↓
& AbsRel↓ & $\delta$<1.25↑ \\
\midrule
Baseline ($\pi^3$*)
& 0.228 & 34.39 / 33.32 & 0.194 & 0.709 \\

+ PanoCity
& 0.256 & 27.49 / 31.52 & 0.105 & \underline{0.896} \\

+ Tri-axis aug
& 0.285 & \textbf{22.05} / 35.04 & 0.116 & 0.871 \\

+ Global pt. fusion
& \underline{0.380} & \underline{24.15} / \underline{26.34} & \underline{0.103} & 0.894 \\

+ Pos. embedding (Complete)
& \textbf{0.427} & 24.53 / \textbf{18.42} & \textbf{0.098} & \textbf{0.908} \\
\bottomrule
\end{tabular}
\vspace{-2mm}
\end{table}

\subsection{Ablation Studies}

Ablation experiments are conducted on Matterport3D using three-view inputs ($N=3$) to quantify the contribution of each component in PanoVGGT. We start from a panoramic adaptation of $\pi^3$ as the baseline and progressively add: (i) PanoCity training data, (ii) tri-axis spherical augmentation, (iii) a global point-fusion head, and (iv) spherical sin–cos positional embeddings. To ensure fair comparison under equal training budgets, all variants are evaluated from the epoch-100 checkpoint, although the complete model is trained for 200 epochs in total. Results are summarized in \cref{tab:ablation}. Notably, the translation error slightly increases after introducing tri-axis spherical augmentation. This occurs because full $SO(3)$ rotations introduce greater orientation diversity, causing early optimization to focus more on rotational alignment. Under the short 100-epoch training schedule used for ablation, this temporarily delays the convergence of translation estimation.

\section{Conclusion}
\label{conclusion}
We presented PanoVGGT, a geometry-grounded Transformer for feed-forward 3D reconstruction from $360^\circ$ imagery. The model jointly predicts camera poses, dense depth, and globally consistent point clouds from unordered panoramic inputs in a single forward pass. By incorporating spherical-aware positional encoding and three-axis $SO(3)$ augmentation, PanoVGGT learns rotation-aware representations that effectively handle severe panoramic projection distortions. We further introduce PanoCity, a large-scale outdoor panoramic dataset with dense depth and pose supervision. Experiments across diverse indoor and outdoor benchmarks demonstrate that PanoVGGT achieves strong reconstruction accuracy and robust geometric reasoning. While PanoVGGT advances scalable panoramic 3D perception, pose estimation remains sensitive to extreme domain shifts and the current design is tailored to equirectangular projections; a detailed discussion of limitations and future directions is provided in the supplementary material.

\section*{Acknowledgment}
\label{acknowledge}
The authors are grateful for the valuable comments and suggestions provided by the reviewers and the Area Chair (AC). This work was supported by the National Natural Science Foundation of China (NSFC) under Grant No. 62406194, the Shanghai Frontiers Science Center of Human-centered Artificial Intelligence (ShangHAI), and the MoE Key Laboratory of Intelligent Perception and Human-Machine Collaboration (KLIP-HuMaCo). We also acknowledge the support from the HPC Platform and the Core Facility Platform of Computer Science and Communication, SIST, ShanghaiTech University, for providing the necessary computational resources and experimental facilities.

{
    \small
    \bibliographystyle{ieeenat_fullname}
    \bibliography{main}
}


\clearpage
\maketitlesupplementary
\appendix

\makeatletter
\@ifundefined{red}{%
  \@ifundefined{textcolor}{%
    \newcommand{\red}[1]{\textbf{[EDIT: }#1\textbf{]}}%
  }{%
    \newcommand{\red}[1]{\textcolor{red}{#1}}%
  }%
}{%
}
\makeatother

\makeatletter
\newcommand{\maybeincludegraphics}[2][]{%
  \IfFileExists{#2}{\includegraphics[#1]{#2}}{%
    \fbox{%
      \parbox[c][.45\linewidth][c]{.95\linewidth}{\centering\small
        Placeholder (file not found):\\\texttt{#2}\\[2pt]
        \red{Replace this box by uploading the figure at the given path.}%
      }%
    }%
  }%
}
\makeatother

\section{Training and Evaluation Details}
\label{sec:supp_train_details}

\paragraph{Training Details.}
We adopt DINOv2 ViT-B/14 as the image encoder~\cite{oquab2023dinov2}, configured with an input height of $336$ pixels (respecting each dataset’s native ERP aspect ratio), a patch size of $14$, and an embedding dimension of $768$. The proposed Geometry Aggregator consists of $24$ alternating-attention blocks. Training is performed on a consolidated panoramic corpus comprising Matterport3D~\cite{chang2017matterport3d}, Stanford2D3D~\cite{armeni2017joint}, Structured3D~\cite{zheng2020structured3d}, and PanoCity, with respective training and validation splits of $4800/2400/2400/16800$ and $480/240/1400/2400$.

A dynamic set-based dataloader is employed, where each training sample randomly selects $2$--$24$ panoramas (validation uses $2$--$12$). 
Color augmentations include random color jittering, grayscale conversion, and gamma adjustments. Geometric augmentations employ physically consistent $SO(3)$ spherical rotations via spherical resampling, ensuring alignment across RGB, depth, and pose modalities.
We use AdamW for optimization with a peak learning rate of $5{\times}10^{-5}$ and a weight decay of $0.05$. Training follows a $5\%$ linear warm-up schedule ($1{\times}10^{-8}\!\rightarrow\!5{\times}10^{-5}$), followed by a $95\%$ cosine decay ($5{\times}10^{-5}\!\rightarrow\!1{\times}10^{-8}$). All remaining hyperparameters follow those reported in the main paper.

For fair comparison, we retrained a panoramic variant of $\pi^3$, denoted $\pi^3{}^{\ast}$, by adapting its geometric head to equirectangular imagery while keeping datasets, augmentations, and optimization settings identical to ours, except that spherical-rotation augmentation was disabled for $\pi^3{}^{\ast}$~\cite{wang2025pi3}. All other baselines rely on official implementations or publicly released pretrained weights under their default configurations (e.g., VGGT~\cite{wang2025vggt}, MoGe~\cite{wang2025moge}, PanDA~\cite{cao2025panda}, UniK3D~\cite{piccinelli2025unik3d}).

\paragraph{Evaluation Details.}
All evaluations use input resolutions of $518{\times}1036$ pixels. 
For the zero-shot depth estimation benchmark on Pano3D~\cite{albanis2021pano3d}, we use single-view inputs, as the dataset provides no cross-view geometric supervision. For 3D point cloud evaluation, we assess multi-view fusion quality by sampling $10$ frames per scene for PanoCity and $3$ frames per scene for both Stanford2D3D~\cite{armeni2017joint} and Matterport3D~\cite{chang2017matterport3d}. The network predicts point maps in both the camera frame and a unified world (anchor) frame; for evaluation, all points are transformed into the world coordinate system using the predicted camera poses. We maintain consistent evaluation protocols across all datasets, using the same voxel size, nearest-neighbor distance metric, and Accuracy/Completeness thresholds as defined in the main paper, with any exceptions noted beneath the respective tables.

We note that Stanford2D3D and Matterport3D were not originally designed for multi-view panoramic reconstruction and therefore contain scenes with limited or negligible overlap between panoramas even after re-splitting and post-processing~\cite{armeni2017joint,chang2017matterport3d}. 
Despite the scarcity of tightly aligned panoramic 3D data, these datasets are retained for training, validation, and evaluation to broaden coverage and to test model robustness under varying levels of viewpoint overlap.


\section{Additional Depth Results}
We extend the depth estimation experiments by adding the retrained panoramic variant $\pi^3$*~\cite{wang2025pi3} to the evaluation. 
All methods follow identical preprocessing, data splits, and evaluation protocols as in the main paper to ensure direct comparability. 
Table~\ref{tab:depth_in_domain_supp} presents the updated monocular results on \emph{Matterport3D}~\cite{chang2017matterport3d}, \emph{Stanford2D3D}~\cite{armeni2017joint}, \emph{Structured3D}~\cite{zheng2020structured3d}, and \emph{PanoCity}. 
Across all datasets, PanoVGGT maintains the best Abs Rel and $\delta{<}1.25$ accuracy, while $\pi^3$*—although retrained on panoramic imagery—remains clearly inferior.
This shows that simply adapting a pinhole-based architecture to panoramic inputs does not yield precise depth estimation, whereas the proposed PanoVGGT effectively learns panoramic geometry from diverse training data.

To further test cross-dataset generalization, we evaluate both models on the Pano3D (GibsonV2) benchmark~\cite{albanis2021pano3d,xia2019gibson} and report the results in Tab.~\ref{tab:supp_zero_shot_pano3d}. 
Under the Scale-only configuration, $\pi^3$* achieves moderate accuracy but remains less stable than PanoVGGT, which attains lower errors across all metrics.
With Scale+Shift alignment, $\pi^3$* improves slightly yet still lags behind in accuracy and structural consistency.
These results confirm that direct retraining on panoramic data provides limited benefit, while our explicit panoramic design maintains robustness under distribution shift.

\begin{table*}[t]
\centering
\scriptsize
\setlength{\tabcolsep}{8pt}
\caption{Monocular depth estimation performance including the retrained panoramic variant $\pi^3$*~\cite{wang2025pi3} on \emph{Matterport3D}~\cite{chang2017matterport3d}, \emph{Stanford2D3D}~\cite{armeni2017joint}, \emph{Structured3D}~\cite{zheng2020structured3d}, and \emph{PanoCity}. 
\textbf{Bold} = best, \underline{underline} = second best.
}
\label{tab:depth_in_domain_supp}
\begin{tabular}{llcccccccccc}
\toprule
Model & Input &
\multicolumn{2}{c}{Matterport3D (Indoor)} &
\multicolumn{2}{c}{Stanford2D3D (Indoor)} &
\multicolumn{2}{c}{Structured3D (Indoor)} &
\multicolumn{2}{c}{PanoCity (Outdoor)} \\
\cmidrule(lr){3-4}\cmidrule(lr){5-6}\cmidrule(lr){7-8}\cmidrule(lr){9-10}
& & Abs Rel $\downarrow$ & $\delta{<}1.25$ $\uparrow$
& Abs Rel $\downarrow$ & $\delta{<}1.25$ $\uparrow$
& Abs Rel $\downarrow$ & $\delta{<}1.25$ $\uparrow$
& Abs Rel $\downarrow$ & $\delta{<}1.25$ $\uparrow$ \\
\midrule
$\pi^3$*~\cite{wang2025pi3} & Monocular 
& 0.0940 & 0.9142 & 0.0852 & 0.9291 & 0.0652 & 0.9649 & 0.0834 & 0.9161 \\
\midrule
PanoVGGT (Ours) & Monocular    & \underline{0.0884} & \underline{0.9157} & \textbf{0.0711} & \textbf{0.9392} & \underline{0.0438} & \underline{0.9728} & \underline{0.0312} & \underline{0.9713}  \\
PanoVGGT (Ours) & Multi-view   & \textbf{0.0840} & \textbf{0.9266} & \underline{0.0778} & \underline{0.9323} & \textbf{0.0400} & \textbf{0.9870} & \textbf{0.0196} & \textbf{0.9812}  \\
\bottomrule
\end{tabular}
\end{table*}

\begin{table}[t]
\centering
\caption{Zero-shot depth results on \emph{Pano3D (GibsonV2)}~\cite{albanis2021pano3d,xia2019gibson} with $\pi^3$* (retrained panoramic variant)~\cite{wang2025pi3}. 
The top part reports results using Scale-only alignment; the bottom part uses Scale+Shift alignment.}
\label{tab:supp_zero_shot_pano3d}
\setlength{\tabcolsep}{6pt}
\resizebox{\columnwidth}{!}{%
\begin{tabular}{lcccc}
\toprule
Method & Abs Rel $\downarrow$ & RMSE $\downarrow$ & $\delta_{1}$ $\uparrow$ & $\delta_{2}$ $\uparrow$ \\
\midrule
$\pi^3$*~\cite{wang2025pi3} & 0.1879 & 0.3743 & 0.9085 & 0.9717 \\
{PanoVGGT (ours)}            & \textbf{0.0869} & \textbf{0.3069} & \textbf{0.9223} & \textbf{0.9859} \\
\midrule
$\pi^3$*~\cite{wang2025pi3} & 0.1777 & 0.3639 & 0.9197 & 0.9759 \\
{PanoVGGT (ours)}            & \textbf{0.0833} & \textbf{0.3015} & \textbf{0.9299} & \textbf{0.9894} \\
\bottomrule
\end{tabular}%
}
\end{table}

\section{Additional Point-Cloud Results}

We extend the point-cloud evaluation to two indoor panoramic datasets, Matterport3D~\cite{chang2017matterport3d} and Stanford2D3D~\cite{armeni2017joint}. 
To thoroughly investigate alternative $360^\circ$ projections and address how perspective-only approaches perform on panoramic data, we introduce an additional baseline denoted as $\pi^3$$^\dagger$. Following the dodecahedral projection protocol of MoGe, we project each equirectangular panorama into 12 overlapping pinhole views. Consequently, a standard 3-panorama input is expanded into 36 individual perspective images and evaluated using the pre-trained $\pi^3$ model. We compare this against our proposed PanoVGGT, the original $\pi^3$, and its re-trained panoramic variant $\pi^3$*.

Each method is evaluated using Accuracy (Acc), Completion (Comp), and their average (Overall). Both mean and median scores are reported, where lower values indicate better performance. Tables~\ref{tab:supp_pc_matterport} and~\ref{tab:supp_pc_2d3d} summarize the quantitative results. While processing dense pinhole splits ($\pi^3$$^\dagger$) improves over the naive $\pi^3$ baseline, it incurs significant computational overhead. Across both datasets, PanoVGGT operating directly on native equirectangular projections achieves the lowest average errors, demonstrating superior geometric accuracy and view-to-view consistency. All experiments are conducted with the same voxel size, nearest-neighbor distance metric, and Acc/Comp thresholds as adopted in the main paper, unless otherwise specified.

\begin{table}[h]
\centering
\small
\caption{Point-cloud results on \emph{Matterport3D}~\cite{chang2017matterport3d}. $\dagger$ denotes the perspective-based model evaluated on panoramas using the dodecahedral projection protocol of MoGe.}
\label{tab:supp_pc_matterport}
\setlength{\tabcolsep}{6pt}
\resizebox{\columnwidth}{!}{%
\begin{tabular}{lcccccc}
\toprule
\multirow{2}{*}{Method} &
\multicolumn{2}{c}{Acc $\downarrow$} &
\multicolumn{2}{c}{Comp $\downarrow$} &
\multicolumn{2}{c}{Overall $\downarrow$} \\
\cmidrule(lr){2-3}\cmidrule(lr){4-5}\cmidrule(lr){6-7}
& Mean & Med & Mean & Med & Mean & Med \\
\midrule
$\pi^3$~\cite{wang2025pi3}                & 0.4027 & 0.3853 & 0.9964 & 0.8803 & 0.6995 & 0.6328 \\
$\pi^3$*~\cite{wang2025pi3}               & 0.2661 & 0.2199 & \underline{0.3351} & 0.2378 & 0.3006 & 0.2288 \\
$\pi^3$$^\dagger$~\cite{wang2025pi3}      & 0.1843 & 0.1420 & 0.2647 & 0.1857 & 0.2245 & 0.1638 \\ \midrule
PanoVGGT (global points)                  & \textbf{0.1743} & \textbf{0.1231} & \textbf{0.1530} & \textbf{0.0890} & \textbf{0.1636} & \textbf{0.1060} \\
PanoVGGT (local points)                   & \underline{0.1750} & \underline{0.1242} & \textbf{0.1530} & \underline{0.0932} & \underline{0.1640} & \underline{0.1087} \\
\bottomrule
\end{tabular}%
}
\end{table}

\begin{table}[h]
\centering
\small
\caption{Point-cloud results on \emph{Stanford2D3D}~\cite{armeni2017joint}. $\dagger$ denotes the perspective-based model evaluated on panoramas using the dodecahedral projection protocol of MoGe.}
\label{tab:supp_pc_2d3d}
\setlength{\tabcolsep}{6pt}
\resizebox{\columnwidth}{!}{%
\begin{tabular}{lcccccc}
\toprule
\multirow{2}{*}{Method} &
\multicolumn{2}{c}{Acc $\downarrow$} &
\multicolumn{2}{c}{Comp $\downarrow$} &
\multicolumn{2}{c}{Overall $\downarrow$} \\
\cmidrule(lr){2-3}\cmidrule(lr){4-5}\cmidrule(lr){6-7}
& Mean & Med & Mean & Med & Mean & Med \\
\midrule
$\pi^3$~\cite{wang2025pi3}                & 0.5087 & 0.4667 & 0.9394 & 0.9103 & 0.7241 & 0.6885 \\
$\pi^3$*~\cite{wang2025pi3}               & 0.2590 & 0.2021 & 0.3143 & \underline{0.2245} & 0.2866 & 0.2133 \\
$\pi^3$$^\dagger$~\cite{wang2025pi3}      & 0.2359 & 0.1942 & 0.3668 & 0.2870 & 0.3013 & 0.2406 \\\midrule
PanoVGGT (global points)                  & \textbf{0.2087} & \textbf{0.1752} & \underline{0.2624} & \textbf{0.1943} & \underline{0.2355} & \textbf{0.1848} \\
PanoVGGT (local points)                   & \underline{0.2109} & \underline{0.1786} & \textbf{0.2590} & \textbf{0.1943} & \textbf{0.2349} & \underline{0.1865} \\
\bottomrule
\end{tabular}%
}
\end{table}

\begin{figure*}[t]
\centering
\maybeincludegraphics[width=\textwidth]{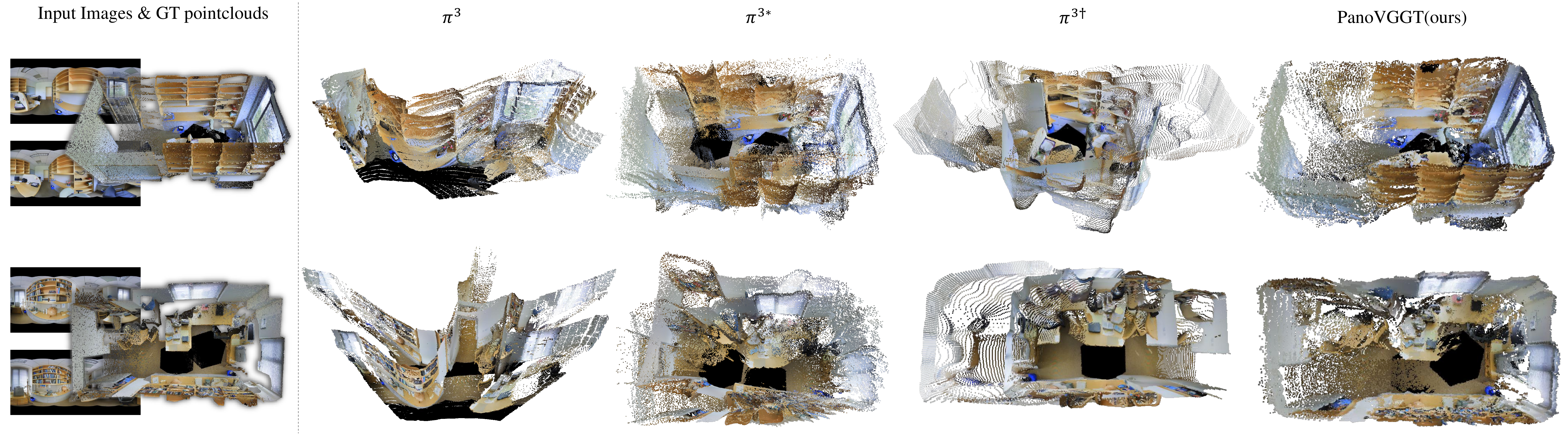}
\caption{Multi-view point-cloud reconstructions on the Stanford2D3D dataset~\cite{armeni2017joint} using two unordered panoramic inputs. 
From left to right: $\pi^3$~\cite{wang2025pi3}, $\pi^3$*~\cite{wang2025pi3}, $\pi^3$$^\dagger$~\cite{wang2025pi3}, and PanoVGGT. 
Our method achieves higher geometric accuracy and cross-view consistency than the baselines.}
\label{fig:supp_vis_stanford}
\end{figure*}

\begin{figure*}[t]
\centering
\maybeincludegraphics[width=\textwidth]{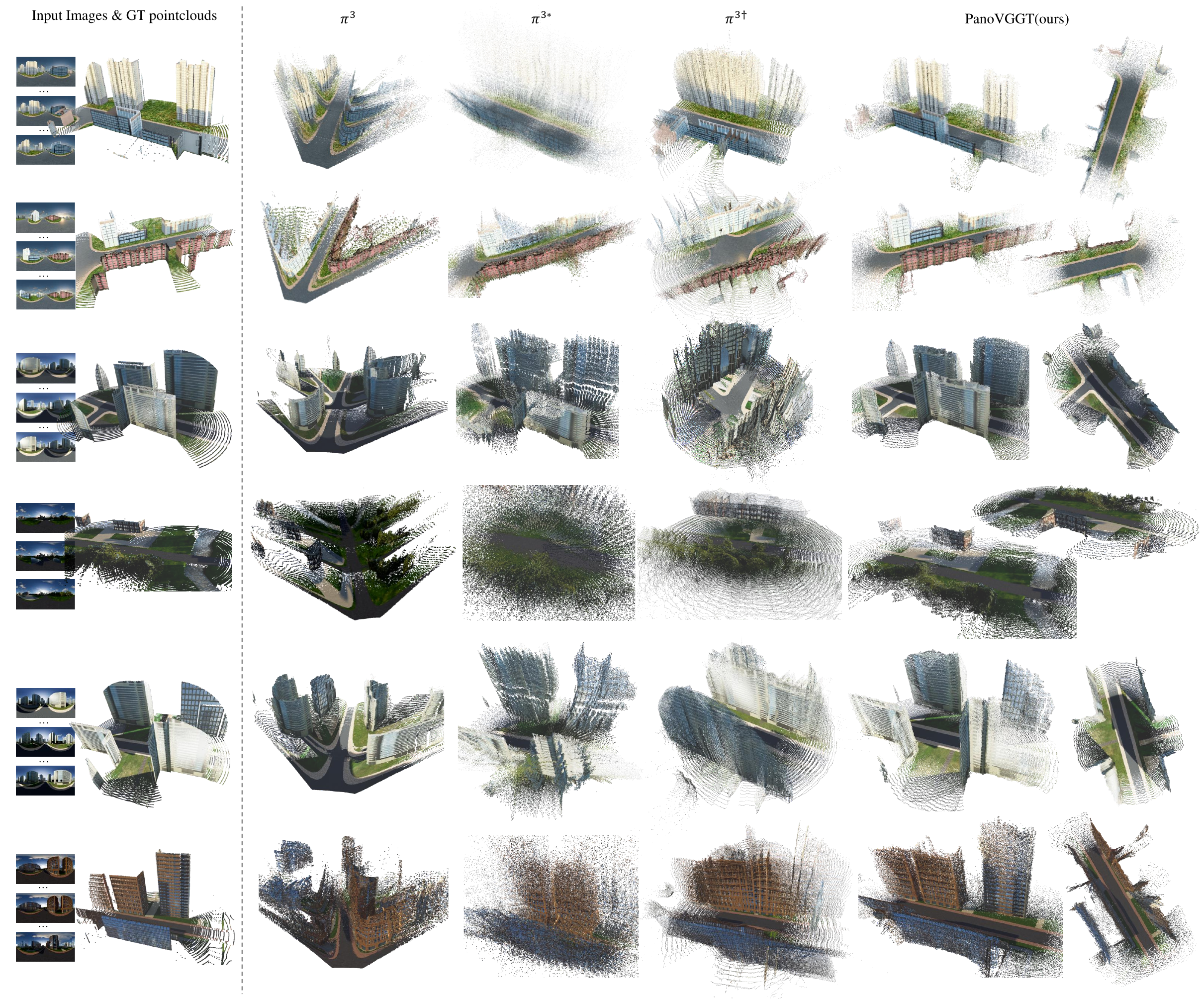}
\caption{Multi-view point-cloud reconstructions on the PanoCity dataset using ten unordered panoramic inputs. 
From left to right: $\pi^3$~\cite{wang2025pi3}, $\pi^3$*~\cite{wang2025pi3}, $\pi^3$$^\dagger$~\cite{wang2025pi3}, and PanoVGGT. 
The baseline methods struggle to learn accurate geometry on this long-trajectory setup, whereas PanoVGGT reconstructs large-scale outdoor scenes with coherent structure and accurate alignment across views.}
\label{fig:supp_vis_panocity}
\end{figure*}

\paragraph{Qualitative Visualizations.}
While qualitative results for Matterport3D~\cite{chang2017matterport3d} are provided in the main paper, \cref{fig:supp_vis_stanford,fig:supp_vis_panocity} show additional multi-view point-cloud reconstructions on Stanford2D3D~\cite{armeni2017joint} and PanoCity. 
Each panel compares the original $\pi^3$~\cite{wang2025pi3}, its retrained panoramic variant $\pi^3$*~\cite{wang2025pi3}, the pinhole-split variant $\pi^3$$^\dagger$~\cite{wang2025pi3}, and the proposed PanoVGGT. 
For the indoor Stanford2D3D dataset, two unordered panoramas are used as input, and reconstructions are displayed from a single viewpoint for clarity, with ceilings removed to expose interior structural details. 
For PanoCity, ten unordered panoramas form a long-trajectory input sequence, and only the PanoVGGT reconstructions are rendered from two viewpoints to illustrate large-scale outdoor geometry and fine structural details. 
These visualizations demonstrate cross-view geometric consistency and enable direct comparison of reconstruction quality across methods.

The original $\pi^3$~\cite{wang2025pi3}, developed for pinhole imagery, directly interprets equirectangular panoramas as pinhole inputs and produces geometrically invalid reconstructions with severe warping and structural distortion. 
The retrained $\pi^3$*~\cite{wang2025pi3} partially adapts but still struggles to produce clean geometry and accurate pose estimates: it often fails to converge on PanoCity and yields incomplete or inconsistent structures on indoor scenes. 
While the dodecahedral projection approach ($\pi^3$$^\dagger$) mitigates severe warping by processing local pinhole views, it introduces visible structural discontinuities among the decoupled patches within each panorama. More severely, it yields inaccurate relative poses between different panoramic views, causing the reconstructed point clouds of the same scene to be misaligned and fail to overlap correctly. 
In contrast, PanoVGGT reconstructs globally consistent 3D scenes with correct geometry and seamless alignment across unordered multi-view panoramas.

\section{Additional Visualization of PanoCity}
\begin{figure*}[t]
\centering
\includegraphics[width=\textwidth]{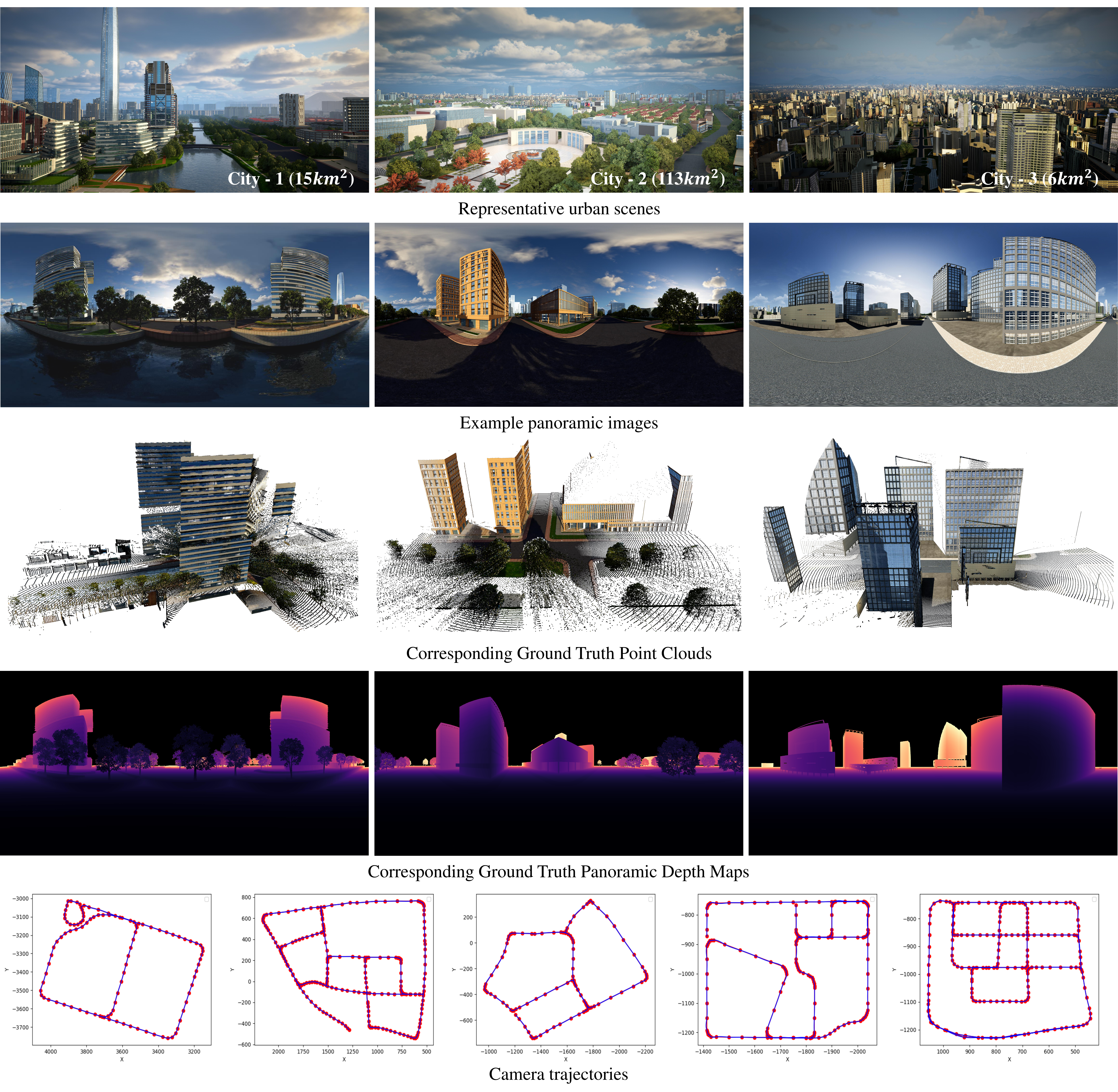}
\caption{Overview of the five core modalities in the PanoCity dataset. 
Row 1: representative urban scenes from three captured cities or regions (City-1, City-2, City-3), annotated with their coverage areas ($15\text{km}^2$, $113\text{km}^2$, $6\text{km}^2$). 
Row 2: example panoramic RGB images collected along the acquisition routes, showing diverse lighting, weather, and façade structures. 
Row 3: corresponding ground-truth 3D point clouds for the same scenes. 
Row 4: ground-truth panoramic depth maps derived from the point clouds, with depth encoded by color. 
Row 5: camera trajectories in bird’s-eye view, illustrating long-range paths with wide baselines and varied viewpoints.}
\label{fig:supp_panocity_overview}
\end{figure*}

~\Cref{fig:supp_panocity_overview} provides an overview of the five key data modalities in the PanoCity dataset, arranged in the order of data acquisition and processing. 
The first row shows representative urban scenes from three different cities or regions (City-1, City-2, City-3), together with their approximate coverage areas ($15\text{km}^2$, $113\text{km}^2$, $6\text{km}^2$). 
These scenes span diverse urban contexts, including dense downtown blocks, residential districts, and major arterial roads. 
The second row presents example equirectangular panoramic RGB images sampled along the acquisition routes, capturing a wide range of lighting conditions, weather variations, and complex architectural structures. 
The third row visualizes the corresponding ground-truth 3D point clouds for these scenes, which provide accurate geometric reference for training and evaluation. 
The fourth row shows the associated ground-truth panoramic depth maps derived from the point clouds; depth is color-coded (e.g., from purple to yellow), making near objects such as trees and roads and far objects such as tall buildings clearly distinguishable in depth. 
The fifth row illustrates the camera trajectories in bird’s-eye view, highlighting long-range paths with large baselines and substantial viewpoint changes. 
Together, these visualizations demonstrate that PanoCity offers large-scale, metrically calibrated panoramic RGB–depth–point-cloud data with realistic urban diversity, making it a suitable testbed for robust panoramic 3D reconstruction and depth estimation.

\end{document}